\pdfoutput=1

\documentclass[11pt]{article}

\usepackage{acl}

\usepackage{times}
\usepackage{latexsym}

\usepackage[T1]{fontenc}

\usepackage[utf8]{inputenc}

\usepackage{microtype}

\usepackage{inconsolata}

\usepackage{graphicx}
\newcommand{\ignore}[1]{}

\usepackage{url}
\usepackage{booktabs}
\usepackage{amsmath}
\usepackage{cleveref}
\usepackage{tikz}
\usepackage{multirow}
\usepackage{amssymb}
\usepackage{tabularx}
\usepackage{pifont}
\usepackage{colortbl}
\usepackage{longtable}
\usepackage[normalem]{ulem}
\usepackage{comment}

\newcommand*\circled[1]{\tikz[baseline=(char.base)]{
            \node[shape=circle,draw,inner sep=.6pt] (char) {#1};}}

\newcommand{\uses}{↑SES}
\newcommand{\lses}{↓SES}

\title{Do Large Language Models Adapt to\\Language Variation across Socioeconomic Status?}

\author{
\textbf{Elisa Bassignana\textsuperscript{1,2}} \hspace{1em} \textbf{Mike Zhang\textsuperscript{3,2}} \hspace{1em} \textbf{Dirk Hovy\textsuperscript{4}} \hspace{1em} \textbf{Amanda Cercas Curry\textsuperscript{5}}\\
 \textsuperscript{1}IT University of Copenhagen \hspace{1em} \textsuperscript{2}Pioneer Center for AI \hspace{1em} \textsuperscript{3}University of Copenhagen \\
 \textsuperscript{4}Bocconi University \hspace{1em} \textsuperscript{5}CENTAI Institute \\
 \texttt{elba@itu.dk}
 }

\begin{document}
\maketitle

\begin{abstract}

Humans adjust their linguistic style to the audience they are addressing. However, the extent to which LLMs adapt to different social contexts is largely unknown. 
As these models increasingly mediate human-to-human communication, their failure to adapt to diverse styles can perpetuate stereotypes and marginalize communities whose linguistic norms are less closely mirrored by the models, thereby reinforcing social stratification.
We study the extent to which LLMs integrate into social media communication across different socioeconomic status (SES) communities.
We collect a novel dataset from Reddit and YouTube, stratified by SES.
We prompt four LLMs with incomplete text from that corpus and compare the LLM-generated completions to the originals along 94 sociolinguistic metrics, including syntactic, rhetorical, and lexical features.
LLMs modulate their style with respect to SES to only a minor extent, often resulting in approximation or caricature, and tend to emulate the style of upper SES more effectively.
Our findings \textit{(1)} show how LLMs risk amplifying linguistic hierarchies and \textit{(2)} call into question their validity for agent-based social simulation, survey experiments, and any research relying on language style as a social signal.

\end{abstract}

\section{Introduction}
Large-scale social media communication directly influences how people use language and how it evolves \cite{AliceEvangalineJebaselvi2023TheIO,dembe2024impact}. From this perspective, social media are not merely a communication tool, but a dynamic environment that actively shapes and transforms language in real-time \cite{akhmedova2024influence}.
Simultaneously, large language models (LLMs) are becoming an integral part of human communication by taking an active role in shaping how users write on social media \cite{anatomyAI,forbes2025ai}, thereby directly influencing language use and communication.

\begin{figure}[t]
    \centering
    \includegraphics[width=\linewidth]{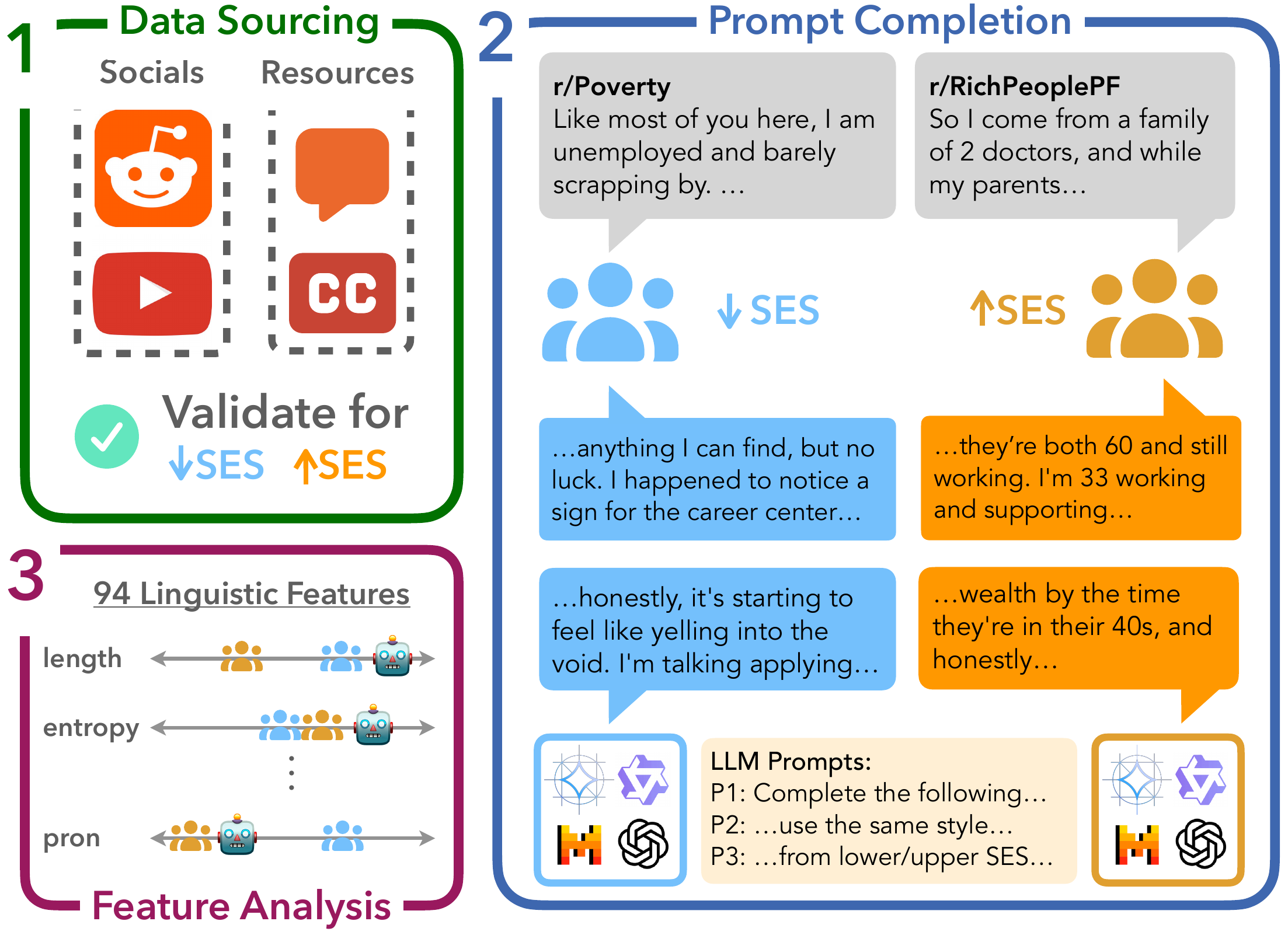}
    \caption{We compare the style of LLM-generated completions against the original text from lower and upper SES communities on Reddit and YouTube along 94 sociolinguistic dimensions.}
    \label{fig:fig1}
\end{figure}

Most work in NLP has focused on analyzing the content of LLM-generated text in terms of factuality \cite{min-etal-2023-factscore,hallucination,wang-etal-2024-factuality} and biases \cite{biasesllms,fang2024bias,gallegos-etal-2024-bias,stranisci-etal-2024-dissecting}. Wrong or biased LLM-generated content has a direct impact on misshaping users' beliefs and behavior \cite{cowritingopinionated,generativeechochamber}.
No work has investigated LLMs' \textit{language style} and its integration into social media communication across different communities. 
However, language style directly and crucially impacts communication \cite{stereotyping,havaldar-etal-2023-comparing,jiang2024impact,li2024influence}, especially large-scale social media communication \cite{mark2011,impactsocialmedia}, and is more indicative of community than topic \cite{tran-ostendorf-2016-characterizing}.

We address this gap by specifically examining the language style of LLM-generated text in large-scale social media communication on Reddit and YouTube. Since style is closely tied to socioeconomic status (SES;~\citealp{Block2020LanguageASA}), we explore the extent to which LLMs emulate the styles of different SES communities when prompted with their texts (see \Cref{fig:fig1}).
Sociolinguistics has shown substantial stylistic differences between individuals from the lower and upper ends of the socioeconomic spectrum \cite{flekova-etal-2016-exploring,curry-etal-2024-classist,bassignana-etal-2025-ai}.
In this work, we address the following research question: 
\textbf{To what extent do LLMs adapt their style to language variation across SES communities in social media communication?}
We adopt a topic-based strategy with keyword matching to systematically collect the first open-access dataset of social media data (Reddit and YouTube), stratified by SES. We validate the stratification of language across SES against previous work \cite{flekova-etal-2016-exploring,basile-etal-2019-write}.
We split each instance (i.e., Reddit posts, YouTube video captions) into two parts and use the first part to prompt four state-of-the-art LLMs with the instruction to complete the text. We then compare the LLM-generated text with the original second part. To formalize the style of LLM-generated text and compare it to in-community language, we employ 94 sociolinguistic metrics that capture syntactic, rhetorical, and lexical features.

Our results show that LLMs only partially match their output to different language styles, for example, in the level of formality marked by the ratio of pronouns and adverbs. 
For most other linguistic features, LLMs' ability to replicate the specific linguistic profile of SES communities varies substantially, often resulting in approximation or caricature rather than precise emulation of the SES in-community language style. 
Our preliminary ablation study on the influence of the input context on model adaptation to SES language variation reveals a latent tendency of LLMs to adapt more effectively to upper SES style, when longer contexts are given.
Our findings contribute to the growing research on how LLMs exacerbate socioeconomic inequalities \cite{bassignana-etal-2025-ai}.
They also have implications for the validity of using LLMs to simulate humans in social science studies \cite{socialsimulacra}.

\paragraph{Contributions.}
\circled{1} We provide the first openly accessible dataset of social media data (Reddit and YouTube) stratified by SES.\footnote{We release our data and code at \url{https://github.com/elisabassignana/language-analysis-social-media}.} \circled{2} We quantitatively evaluate LLMs' ability to replicate SES-specific style across two domains, four LLMs, three prompting strategies, and 94 sociolinguistic features.

\section{Related Work}

\paragraph{Language Variation across SES.}
Since established work in sociolinguistics highlighted the differences in language use by individuals with different socioeconomic statuses \cite{Labov_2006}, several works in NLP have attempted to quantify these differences on a large scale. In the context of social media, \citet{preotiuc-pietro-etal-2015-analysis,preoctiuc2015studying} introduced a dataset of Twitter users mapped to their income through their occupational class using job-related keywords to match user profile fields. \citet{flekova-etal-2016-exploring} employed that dataset to analyze the relation between language use and income on Twitter. 
However, more than for any other widely spread social media, the Twitter community is notoriously skewed towards the upper end of the socioeconomic spectrum \cite{whotweet,wojcik2019sizing}. 
\citet{basile-etal-2019-write} relied on the findings on Twitter by \citet{flekova-etal-2016-exploring} to expand research on stylistic variation as a predictor of social classification in the domain of restaurant reviews.

\paragraph{Language Style of LLM-generated text.}

\citet{MuozOrtiz2023ContrastingLPA} compared the language style of humans against LLM‑generated news, finding measurable differences at the level of grammar, vocabulary, constituents, and dependency distances. \citet{zamaraeva-etal-2025-comparing} expand the analysis of human-written versus LLM-generated NYT-style text through the lens of grammatical structure.
\citet{reihartlikehumans} investigated the ability of LLMs to match human language style in different domains (academic, news, fiction, podcasts, blogs, television, and movie scripts) and identified systematic differences across the style of distinct LLMs.

However, the extent to which LLMs integrate within large-scale communication on social media within different SES communities is largely unknown.
In this work, we expand previous SES-stratified datasets to two new domains (i.e., Reddit and YouTube) and validate our data collection strategy against the findings of previous work.\footnote{We contacted \citet{flekova-etal-2016-exploring} and \citet{basile-etal-2019-write} to request their datasets, but were unsuccessful.} Then, we use our dataset as an anchor to evaluate the language style of LLM-generated text against lower and upper SES communities along 94 features.

\begin{table}[t]
    \centering
    \small
    \begin{tabular}{l|rr|rrrrr}
    \toprule
    & \multicolumn{2}{c|}{\textbf{YouTube}} & \multicolumn{2}{c}{\textbf{Reddit}} \\
    & Lower & Upper & Lower & Upper \\
    \midrule
    \# instances    & 472   & 622       & 1,988       & 1,689    \\
    \# tokens       & 1.06M  & 1.13M      & 315.4K      & 227.8K \\
    \bottomrule
    \end{tabular}
    \caption{\textbf{Statistics} of the lower and upper SES datasets. The \# of instances refers to the number of video captions for YouTube and to the number of posts for Reddit.}
    \label{tab:dataset-statistics}
\end{table}

\section{Data}

To explore the extent to which LLMs emulate SES in-community language on social media, we collect data from Reddit\footnote{We use the Pushshift API dumps and search for posts and comments from 2008 until 2024.} and from YouTube,\footnote{YouTube data collected using the YouTube Data API.} each serving as a platform where distinct SES communities spontaneously congregate.
Although SES communities naturally differ in the topics they discuss—reflecting distinct needs, lifestyles, social circles, and access to experiences—our analysis instead focuses on structural linguistic features, such as style, register, and other sociolinguistic markers, which are not affected by semantic variation. 
\Cref{tab:dataset-statistics} reports the statistics of our dataset, and below we describe our data collection strategy.

\subsection{Lower SES}
\label{sec:lowerses}

Money matters are frequently discussed among lower SES communities \cite{lareau2018unequal,mccaslin2022lifelong,Ndou_2024}. 
Our data collection for the lower SES centers around topics related to \textit{financial struggle}, \textit{poverty}, \textit{frugality}, and \textit{benefits}. To identify relevant social media content, we manually compile a list of keywords related to these themes (e.g., `poverty', `poor', `frugal') and use them to systematically search for subreddits (e.g., r/povertyfinance, r/Frugal) and YouTube videos.\footnote{List of subreddits and YouTube searches in \Cref{app:data}.}
Then, to clean our data and to exclude onlookers from in-groups, we differentiate the process between Reddit and YouTube as follows:
\paragraph{Reddit.} 
We perform a network analysis and clean our data by:
\textit{(1)} filtering only the posts by users that interacted (i.e., posting, commenting) at least 10 times within our set of subreddits and \textit{(2)} interacted in at least three different subreddits within our list.
We remove all usernames ending with `bot', which is how bots are commonly referred to on Reddit (e.g., `u/sneakpeekbot'), and we manually check usernames including `mod', which is often used to refer to automatic moderators. Finally, we omit posts including URLs and select a maximum of one post for each remaining user.

\paragraph{YouTube.} 
As our goal is to collect videos spoken in the first person by individuals belonging to our target group, we query the API using the combination of `vlog' and a keyword. To further clean the retrieved list of videos from non relevant content (e.g., documentaries) we filter in only the videos where the title is written in first person by checking for first-person pronouns (e.g., `I', `we', `my', `us'). Then we use the YouTubeTranscriptApi to scrape the caption of our final set of videos.

To support a more robust analysis of the language style, we retain only instances (Reddit posts and YouTube captions) with at least 50 words. 

\subsection{Upper SES}
\label{sec:upperses}

On the other hand, online communities gathering around subreddits like r/Rich or r/millionaire mainly attract individuals seeking advice on how to achieve a higher level of wealth. In fact, upper SES individuals rarely explicitly discuss their own wealth \cite{10.1162/daed_a_01752}.
Instead, hobbies have been shown to be a highly distinctive dimension for upper SES and a symbol of social identity~\cite{bourdieu_distinction._1984}. We center our data collection for the upper SES around lifestyle, hobbies, and leisure activities. 
\citet{Sawert2021WithinTrackDA} analyzes how upper-class families use specific sports and classical music to transmit privilege across generations.
Based on previous literature \cite{Engstrom,Hwang2012,post2018socioeconomic,eime2015participation,Friedman,arnold2022social,Schmitt,Sawert2021WithinTrackDA,cuijuan2023research}, we collect a list of hobbies and leisure activities (e.g., golf, sailing) that are distinctive of the upper SES and we employ it for systematic data collection on Reddit and YouTube.
Then, we use the same cleaning process as for the lower SES (see \Cref{sec:lowerses}).

\subsection{Dataset Validation}

While collecting data from social media is mostly straightforward, inferring users' SES is a more complex matter. As detailed in \Cref{sec:lowerses,sec:upperses}, we identified SES-based communities from previous literature about SES topics of interest. Here, we externally validate our dataset to further ensure the soundness of our strategy.
\citet{flekova-etal-2016-exploring} showed how the readability of text correlates with income.
Their findings are consistent with observations that readability correlates with education \cite{davenport2014readability}, which has an important role in determining SES \cite{bourdieu2018distinction}. 
Following \citet{basile-etal-2019-write} we validate our dataset against \citet{flekova-etal-2016-exploring}' results on readability metrics: Automated Readability Index (ARI; \citealp{smith1967automated}), Coleman Liau Index \cite{coleman1975computer}, Dale Chall Readability Score \cite{dale1948formula}, Flesch-Kincaid Grade \cite{kincaid1975derivation}, Flesch-Reading Ease \cite{flesch1948new}, Gunning Fog \cite{gunning1952technique} and Linsear Write Formula \cite{usaf1975readability}.\footnote{We use the implementations of \texttt{textstats}: \url{https://github.com/textstat/textstat}.}
Readability metrics are designed to estimate a text's complexity, typically by analyzing the average number of syllables per word and words per sentence.
Following \citet{flekova-etal-2016-exploring} and \citet{basile-etal-2019-write}, we expect the readability scores to increase from the lower SES to upper SES subsets, except for Flesch-Reading Ease, which, by definition of the metric, leads us to expect an inverse correlation.
\Cref{tab:readability} reports the readability metrics on our dataset, where the scores follow the expected trends.\footnote{Readability metrics are designed for written language. Therefore, the metrics in \Cref{tab:readability} refer to the Reddit dataset. For completeness, we also report the readability scores of the YouTube portion of the dataset in \Cref{app:readabilityYT}.} The differences are statistically significant according to a Mann-Whitney U test~\citep{mann1947test}. Similar to~\citet{basile-etal-2019-write}, only the Linsear Write Formula does not show a significant difference across SES groups.

\begin{table}[]
    \centering
    \small
    \begin{tabular}{l|rr}
    \toprule
    \textbf{Metric} & \textbf{Lower SES} & \textbf{Upper SES}  \\
    \midrule
     ARI\textsuperscript{$*$}             & 7.18 & 7.36 \\
     Coleman-Liau\textsuperscript{$*$}    & 6.16 & 6.43 \\
     Dale-Chall\textsuperscript{$*$}      & 8.45 & 8.76 \\
     Flesch-Kincaid\textsuperscript{$*$}  & 7.04 & 7.18 \\
     Flesch-Reading\textsuperscript{$*$}  & 73.99 & 72.45  \\
     Gunning-Fog\textsuperscript{$*$}     & 9.12 & 9.29\\
     Linsear                              & 8.81 & 8.88  \\
    \bottomrule
    \end{tabular}
    \caption{\textbf{Mean Readability Scores per SES Group (Reddit).} (\textsuperscript{$*$}) indicates a statistically significant difference in the distributions of readability scores ($p < 0.05$) between lower and upper SES, as determined by a Mann-Whitney U test~\citep{mann1947test}.}
    \label{tab:readability}
\end{table}

\section{LLM-Generated Data}
\label{sec:prompts}

To analyze whether and to what extent LLMs emulate in-community language style based on the input prompt, we generate social media data starting from the data we collected.
Following \citet{reihartlikehumans} we split each instance in our dataset (Reddit posts and YouTube captions) in two parts. We use the first part (i.e., 25 words) as a language cue for the input prompt and instruct the models to complete the text. We use the second part (i.e., the remaining text) as a comparison with the LLM-generated content. 
We analyze three variations of the prompt, which increasingly explicit the models to adapt their output with respect to the input: 
\begin{enumerate}
    \itemsep0em
    \item \textbf{Implicit (IMP):} ``Complete the following [\textit{Reddit post / caption for a YouTube video}]. Only generate the completion and nothing else. \verb|\n{text}|'';
    \item \textbf{Explicit Language Style (ELS):} ``Complete the following [\textit{Reddit post / caption for a YouTube video}] using the same style, tone, and diction of the first part. Only generate the completion and nothing else. \verb|\n{text}|'';
    \item \textbf{Explicit Language Style + SES (ELS-SES):} ``Complete the following [\textit{Reddit post / caption for a YouTube video]} written by a user from a [\textit{lower / upper}] socioeconomic status using the same style, tone, and diction of the first part. Only generate the completion and nothing else. \verb|\n{text}|''.
\end{enumerate}
We experiment with Gemma-3-27B-it~\cite{gemmateam2025gemma3technicalreport}, Mistral-Small-3.2-24B-Instruct-2506~\citep{mistral2025small}, Qwen3-30B-A3B-Instruct-2507~\citep{qwen3technicalreport}, and GPT-5~\citep{openai2025gpt5}. For inference, we generate one chat completion per prompt with the default temperature. Finally, we depict the hardware, inference costs and environmental impact in~\Cref{app:inference}.

\section{Linguistic Analysis}

We rely on 94 sociolinguistic metrics to analyze the style of lower and upper SES communities, as well as the style of the LLM-generated text. 

\paragraph{Biber features.} We use Douglas Biber's set of 67 linguistic categories collected from \citet{biber1991variation,biber1995dimensions}, and normalize them by the instance length. The set includes lexical, grammatical, and rhetorical features (e.g., pronouns, tense, place and time adverbials, discourse particles, adjectives before a noun). We report the full list in \Cref{app:biber}.\footnote{We rely on the implementation of the pseudobibeR package: \url{https://github.com/browndw/pseudobibeR}.}

\paragraph{Part of Speech.}
Texts with more nouns and articles as opposed to pronouns and adverbs are considered more formal~\cite{pennebaker2003psychological,Argamon2009,rangel2014overview}. We use Spacy's POS-tagger and normalize each count by the length of the instance.

\paragraph{Length.}
As shorter words are considered more readable \cite{gunning1969fog,pitler-nenkova-2008-revisiting}, and following \citet{flekova-etal-2016-exploring}, we calculate the number of words, syllables, lexicon, sentences, characters, letters, polysyllables, and monosyllables.

\paragraph{Style.} 
We assess the level of concreteness using the list proposed by \citet{Brysbaert2014ConcretenessRF}. In this collection, each word is labeled on a scale from one (abstract) to five (concrete). We compute the level of concreteness as the mean value of the words in a text. Additionally, we compute the entropy and utilize Spacy to determine the maximum depth of the syntax dependency trees as a measure of syntactic complexity. Last, following \citet{curry-etal-2024-classist} we compute the ratio of named entities (NE), and the ratio of hapax legomena.

\begin{figure*}
    \centering
    \includegraphics[width=\linewidth]{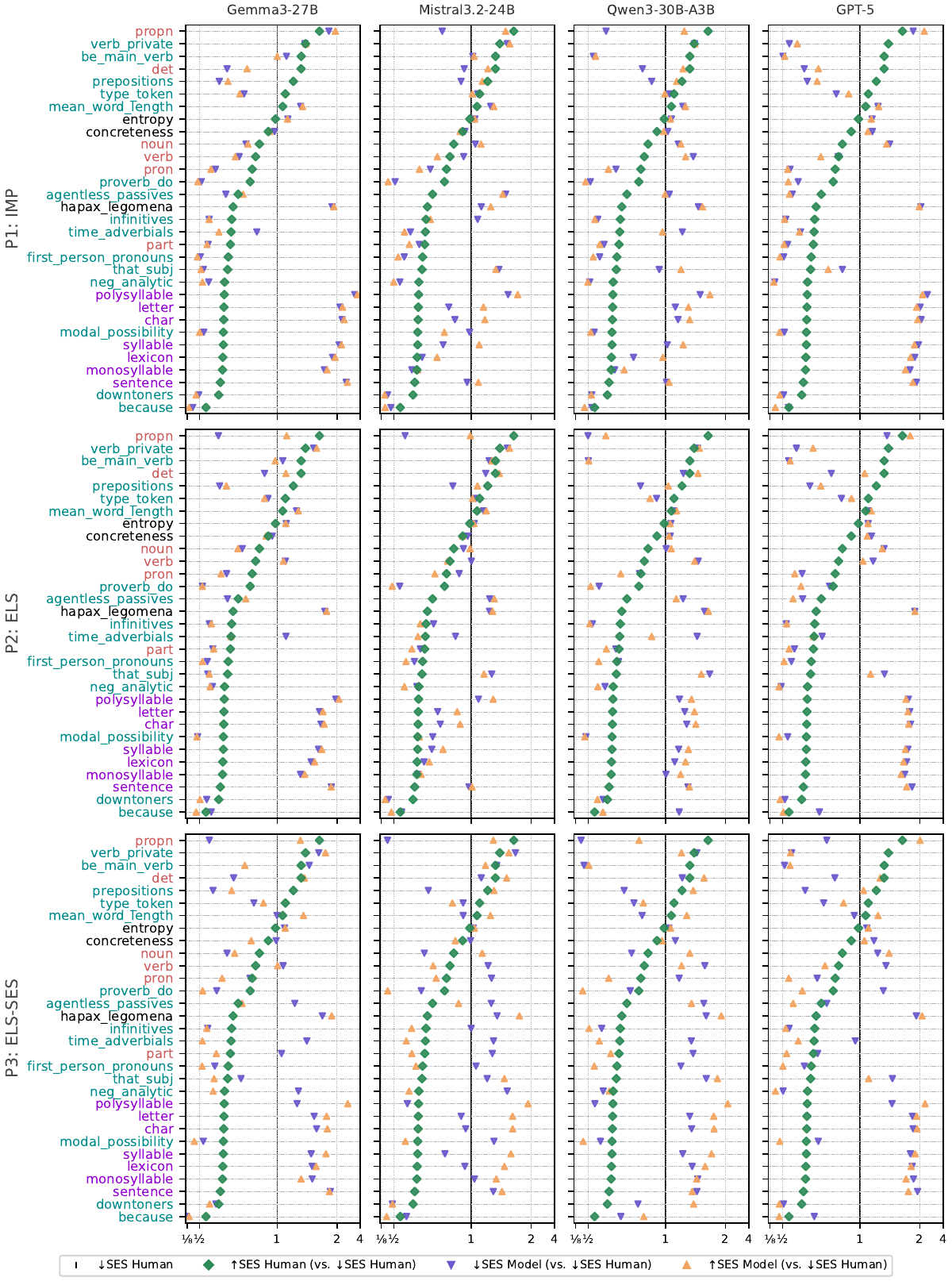}
    \caption{\textbf{Forest Plots Comparing Linguistic Features of Humans and Models on Reddit.} We \textit{only} show the linguistic features (31) with a statistically significant difference with correction ($p<0.01$;~\citealp{mann1947test, holm1979simple}) in usage between lower SES (↓SES; rate $= 1$) and upper SES (↑SES) human writers. Each point indicates the frequency ratio of a feature in a model's (or human's) output compared to human text from the \lses{} group. These comparisons are presented across four models and three prompts (see Section~\ref{sec:prompts}). Feature types are color-coded: Biber features are cyan, length-specific features are dark violet, PoS tags are red, and style features are black.} 
    \label{fig:reddit-all}
\end{figure*}

\begin{figure*}[t]
    \centering
    \includegraphics[width=\linewidth]{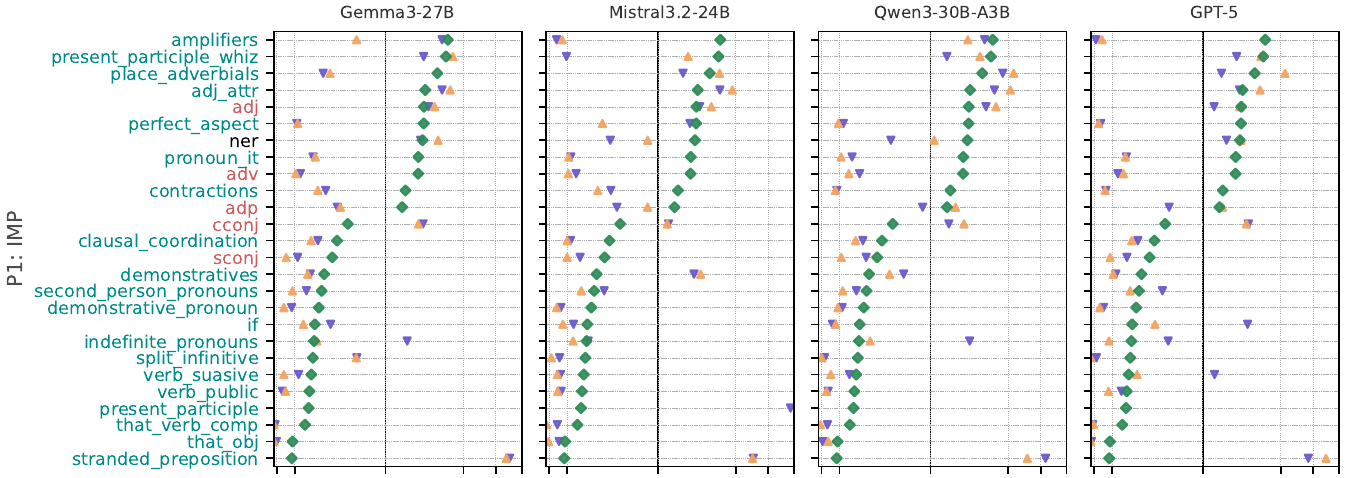}
    \includegraphics[width=\linewidth]{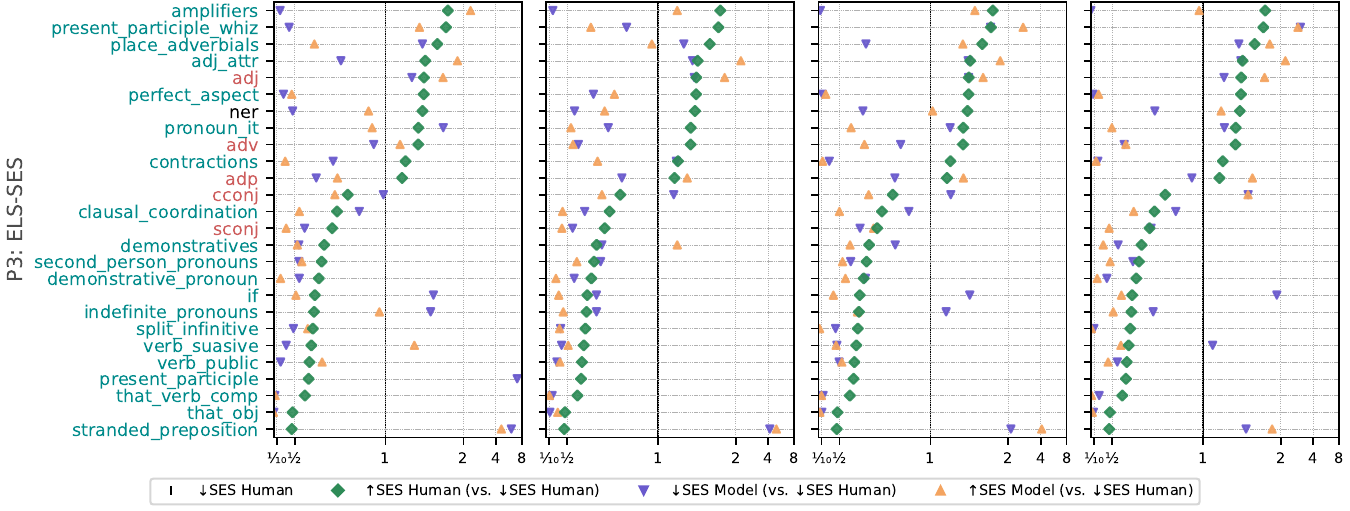}
    \caption{\textbf{Comparison of Linguistic Features on YouTube.} 
    This plot displays \textit{only} the linguistic features, not present in Reddit results, that show a statistically significant difference ($p<0.01$, corrected) between lower SES (↓SES) and upper SES (↑SES) human writers. Each point represents the frequency ratio of a feature relative to the ↓SES human group. Feature types are color-coded: Biber (cyan), length (dark violet), PoS (red), and style (black).}
    \label{fig:yt-exclusion}
\end{figure*}

\section{Results}

\subsection{Reddit}

In~\Cref{fig:reddit-all}, we show our results on Reddit data across three prompts of increasingly explicit language style and four state-of-the-art open and closed-source models (\Cref{sec:prompts}). 
We report features that are statistically significantly different according to a Mann-Whitney U test~\citep{mann1947test} between lower SES (\lses{}) and upper SES (\uses{}) online communities, with a Holm-Bonferroni correction~\citep{holm1979simple} applied to control for multiple comparisons at an $\alpha = 0.01$.
We report the complete results in \Cref{app:extended-results}. To facilitate visualization, in the forest plot in~\Cref{fig:reddit-all}, we use lower SES humans as an anchor and compare the frequency ratio of the features against it.

\paragraph{Biber features.}

Individuals from upper SES communities tend to have a wider and more sophisticated vocabulary \cite{du2022collocation}. These differences emerge in our results through higher values of `mean\_word\_length' and token type ratio for upper SES over lower SES communities. In respect, LLMs generate sequences with similar word length for both communities, except when prompt 3 (ELS-SES) makes the SES group explicit:  In this case, all four LLMs exacerbate the differences by generating even shorter (longer) words than the corresponding lower (upper) SES communities. The token type ratio of generated text is always lower than that of human-written text, indicating a more repetitive use of language.
The use of `that' as a subject indicates simpler and more colloquial syntactic structures, which are more frequent among lower SES. LLMs generally generate this structure with a higher frequency with respect to the corresponding human-written text (except for Gemma). Trends across the lower and upper SES generated texts are mixed across prompt types for all four models, indicating that LLMs do not pick up on the use of `that' as a subject.

\paragraph{Part of Speech.}

The higher rate of determiners (`det') in the language used by the upper SES is largely reflected in the output of all four LLMs, with prompt 3 exacerbating this gap. The different rates of nouns in lower and upper SES language are not substantial in the corresponding lower and upper SES-generated text, except for the output of Mistral and Qwen for prompt 3. In this case, however, both models generate an opposed trend with respect to the lower/upper human-written text. Finally, in-community language contains a higher rate of verbs, pronouns (including `pron' and `first\_person\_pronouns'), and adverbs (`time\_adverbials', `downtoners') for lower SES \cite{jones1973speech,shi2021lexical}, indicating a more informal style \cite{pennebaker2003psychological,Argamon2009,rangel2014overview}. The ratio of these PoS in the LLM-generated text follows the respective lower and upper SES communities trend, with exacerbated differences when including the SES in the prompt (prompt 3).

\paragraph{Length.}

LLMs have a general bias towards verbosity \cite{Saito2023VerbosityBIA}. Consistently, in our experiments, LLM-generated text is generally longer than human-written text. Especially Gemma and GPT generate text that is approximately twice as long as the corresponding human version, for all three prompt types. Mistral tends to generate shorter sequences of text and, similarly to Qwen, to differentiate more between lower and upper SES. However, both models generate longer sequences for upper SES and shorter sequences for lower SES (especially when using prompt 3), which contrasts with the trend observed in lower/upper communities. In fact, individuals with upper SES tend to express themselves with fewer words, likely due to a wider vocabulary \cite{bassignana-etal-2025-ai}.

\paragraph{Style.}

\citet{bernstein-language} posits that individuals from upper SES families are more encouraged to use language for abstract thinking in contrast to people from lower SES families, who are exposed to more concrete concepts. This difference is reflected in the language style adopted by lower and upper-SES communities on social media. LLMs pick up on the different levels of concreteness within various communities to a limited extent, which is further amplified in the results obtained with prompt 3. The hapax legomena is higher for lower SES communities (as well as to a smaller extent, the entropy) that we speculate being a consequence of lower SES language containing more non-standard expressions and slang with respect to the upper SES counterpart \cite{LANSLEY201685}. LLM-generated text does not capture these differences across SES communities (prompts 1 and 2). When making SES explicit (prompt 3), Mistral and Qwen differentiate the text; however, the results trend in the opposite direction compared to human values (i.e., lower values of hapax legomena for the completion of the lower SES prompts).

Overall, we report little differences between the results obtained with the first and the second prompt strategies (IMP, ELS), indicating that LLMs do not easily pick up the language cues of the input text, even when explicitly prompted to replicate the style. When the input prompt contains explicit information about the social community to emulate (lower or upper SES), LLMs differentiate the generated output more distinctly. However, language differences are often exaggerated with respect to human values or do not accurately reflect the real trends across SES communities.

\subsection{YouTube}

In~\Cref{fig:yt-exclusion}, we show our results on YouTube data for prompts 1 and 3 (IMP and ELS-SES), focusing on the features that are distinctive of spoken language (i.e., that are not statistically significantly different in Reddit as well). We report the complete results in \Cref{app:extended-results}. 
As for Reddit, LLMs generally struggle to match the right register in terms of formality. The results for informality markers, such as contractions and stranded prepositions, show inverse trends with respect to both lower- and upper-SES communities. 
The language of upper SES communities is characterized by higher syntax complexity: LLMs correctly identify the association with present participial postnominal (`present\_participle\_whiz') and increase their usage when generating upper SES text (except Mistral and GPT for prompt 3), but under-produce perfect aspect.
All four models tend to exhibit the correct lower/upper SES trends in terms of indefinite pronouns, with less frequent usage in upper SES communities, although with varying proportions compared to human language. 
Instead, they undergenerate demonstrative pronouns and fail to differentiate across SES community trends.

Similar to the Reddit results, all models demonstrate some ability to modulate their style, particularly with the explicit ELS-SES prompt. However, their ability to accurately replicate the specific linguistic profile of a language community varies substantially, often resulting in approximation or caricature rather than precise emulation of the in-community language style.

\begin{figure}[t]
    \centering
    \includegraphics[width=\linewidth]{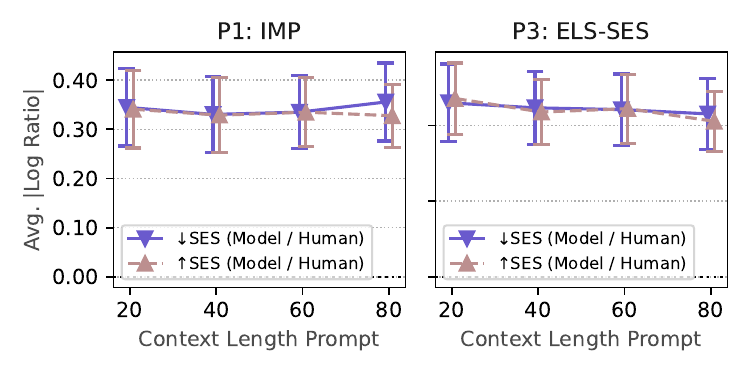}
    \caption{Average absolute logarithm of the ratio between model and human text across increasing context.}
    \label{fig:ablation}
\end{figure}

\section{Ablation Study on the Input Context}

We investigate the impact of context length on LLMs' tendency to pick up on linguistic cues from the input prompt and modulate the language style of the output accordingly.
We use the same methodology presented in \Cref{sec:prompts}, but we experiment with increasingly long input contexts.
We filter the instances with at least 100 words and split them using, respectively, the first 20, 40, 60, 80 words for prompting and the rest for the comparison with the LLM-generated text.
For this investigation, due to computational constraints, we focus on Reddit, leave out prompt 2 (ELS), and experiment with Gemma only.
\Cref{fig:ablation} presents the aggregated results, showing the average absolute logarithmic ratio between human and LLM text for lower and upper SES, respectively, across all features.\footnote{We leave out of this analysis the features related to the length to focus on the style.} We observe: \textit{(1)} 
Only minor variation across increasing context length (and mainly for prompt 3), indicating that LLMs rely on linguistic cues from the input prompt to modulate their output for different language styles only to a minor extent. \textit{(2)}
When increasing the context to 80 words, the ratio between human and LLM is smaller for upper SES, indicating a latent tendency of LLMs (Gemma) to integrate better within upper SES language style.

\section{Discussion and Conclusion}

Language style is closely tied to SES \cite{Block2020LanguageASA}.
We investigate the extent to which LLMs adapt their style to language variation across lower and upper SES communities. 
We compare the stylistic variation of online SES communities against LLM-generated text.
While the language adopted by lower SES is less formal, less predictable due to the jargon adopted, and more concrete, 
LLMs modulate their style to SES in-community language only to a limited extent, often resulting in approximation or caricature rather than accurate emulation. 
A longer context does not facilitate LLMs in capturing the style of different language communities. Notably, our ablation study indicates a latent tendency for models to better emulate the style of upper SES communities compared to lower SES. 

Our findings reveal severe consequences for the widespread adoption of LLMs for communication: if LLMs modulate their style more easily towards certain communities than others, they directly contribute to exacerbating social inequalities \cite{capraro2024impact}. Communities that are not accurately represented may experience a degraded user experience, hindering the adoption of these technologies \cite{davis1989perceived}, and ultimately contributing to the exacerbation of the so-called `AI-gap' \cite{bassignana-etal-2025-ai}.
Additionally, our results pose a new challenge to the validity of using LLMs to emulate humans for agent-based social simulation and research relying on language style as a social signal \cite{Argyle2022OutOOA,Aher2022UsingLLA}.

\section*{Limitations}
While our study reveals important disparities in how LLMs model sociolinguistic variation, it is limited in terms of:
\begin{itemize}
    \item \textbf{Dataset}: Our analysis focuses on two social media platforms. Language patterns differ based on the affordances and specific communities of each platform. 
    
    \item \textbf{Operationalisation of SES}: As previous work on NLP and SES, our analysis uses proxies rather than self-reported or objective measures of SES. In addition, we do not consider a middle-SES. 
    
    \item \textbf{Choice and scope of linguistic measures}: We used a set of 94 surface metrics supported from previous studies in sociolinguistics; however, it is not exhaustive, and it is possible they might miss something more nuanced. 
    
    \item \textbf{Prompt design}: LLM outputs are strongly dependent on the prompt used to generate them. While we test three different prompt variations, it is possible that a different prompt or prompting strategy would result in stronger variation. 
    
    \item \textbf{Models}: We tested a limited number of models, and new models are published regularly. It is possible that our findings may not hold across future model versions.
\end{itemize}

\paragraph{}

\section*{Ethical Considerations}

For the analysis proposed in the paper, we utilize Reddit data from the Pushshift API dumps and YouTube data scraped using the YouTube Data API, adhering to the developer guidelines of the respective platforms. We do not collect any user-identifiable information (i.e., Reddit and YouTube usernames).

\section*{Acknowledgments}

We thank Francesco Corso for his help in collecting and cleaning the Reddit data.
We thank the MilaNLP group at Bocconi University for feedback on an earlier version of this draft.
Elisa Bassignana is supported by a research grant (VIL59826) from VILLUM FONDEN.
Mike Zhang is supported by funding from the Danish Government to Danish Foundation Models (4378-00001B).
Dirk Hovy is supported by the European Research Council (ERC) under the European Union’s Horizon 2020 research and innovation program (grant agreement No.\ 949944, INTEGRATOR). He is a member the Data and Marketing Insights Unit of the Bocconi Institute for Data Science and Analysis (BIDSA).

\bibliography{anthology,custom}

\appendix

\section{Data Collection}
\label{app:data}

\paragraph{Lower SES.}
\begin{itemize}
    \item Subreddits list: povertyfinance, povertykitchen, frugalmalefashion, FrugalFemaleFashion, FrugalLiving, TrueFrugal, FrugalShopping, poor, Frugal, Cheap\_Meals.
    \item YouTube searches: low income, living on minimum wage, surviving on food stamps, food stamps experience, living paycheck to paycheck, budgeting on minimum wage, food bank haul, food pantry haul, public housing tour, homeless, unemployed, section 8 apartment tour, feeding a family on a budget, living in poverty, eviction.
\end{itemize}

\paragraph{Upper SES.}

\begin{itemize}
    \item Subreddits list: RichPeoplePF, fatFIRE, Philanthropy, golf, Rowing, horseracing, Equestrian, Horses, tennis, 10s, Shooting, CompetitionShooting, Hunting, polo, sailing, opera, classicalmusic, FineArt, FineArtPhoto, literature, Fencing, Fieldhockey, yachting.
    \item YouTube searches: golf, rowing, equestrian, tennis, horse riding, sailing, fine arts, opera, classical music, literature, fencing, field hockey, yachting, yacht.
\end{itemize}

\section{Readability Metrics YouTube}
\label{app:readabilityYT}

Readability metrics are designed for written language, which is in principle different from spoken language \cite{oralandliterate,ortmann-dipper-2019-variation}. However, for completeness, here we report the readability metrics computed on the YouTube portion of our dataset. As these metrics typically normalize by sentence length, and YouTube transcripts are not split into sentences, we first run the \texttt{wtpsplit} sentence tokenizer.\footnote{\url{https://github.com/segment-any-text/wtpsplit/}} Differently from Reddit, the metrics computed on YouTube captions do not differ significantly between SES communities (except for Flesch-Reading).

\begin{table}[]
    \centering
    \small
    \begin{tabular}{l|rr}
    \toprule
    \textbf{Metric} & \textbf{Lower SES} & \textbf{Upper SES}  \\
    \midrule
     ARI             & 4.03 & 3.78 \\
     Coleman-Liau    & 4.19 & 4.17 \\
     Dale-Chall      & 7.19 & 7.28 \\
     Flesch-Kincaid  & 4.61 & 4.21 \\
     Flesch-Reading\textsuperscript{$*$}  & 85.01 & 85.71  \\
     Gunning-Fog     & 6.81 & 6.30 \\
     Linsear                              & 6.73 & 5.87  \\
    \bottomrule
    \end{tabular}
    \caption{\textbf{Mean Readability Scores per SES Group (YouTube).} (\textsuperscript{$*$}) indicates a statistically significant difference in the distributions of readability scores ($p < 0.05$) between lower and upper SES, as determined by a Mann-Whitney U test~\citep{mann1947test}.}
    \label{tab:readability-yt}
\end{table}

\section{Biber's Features}
\label{app:biber}

Below we list the full set of Biber's features adapted from \citet{biber1991variation,biber1995dimensions} that we compute using the pseudobibeR library.\footnote{\url{https://github.com/browndw/pseudobibeR}.}

\paragraph{Tense and aspect markers}
\begin{itemize}
    \item \texttt{past\_tense}: Past tense
    \item \texttt{perfect\_aspect}: Perfect aspect
    \item \texttt{present\_tense}: Present tense
\end{itemize}

\paragraph{Place and time adverbials}
\begin{itemize}
    \item \texttt{place\_adverbials}: Place adverbials (e.g., above, beside, outdoors)
    \item \texttt{time\_adverbials}: Time adverbials (e.g., early, instantly, soon)
\end{itemize}

\paragraph{Pronouns and pro-verbs}
\begin{itemize}
    \item \texttt{first\_person\_pronouns}: First-person pronouns
    \item \texttt{second\_person\_pronouns}: Second-person pronouns
    \item \texttt{third\_person\_pronouns}: Third-person personal pronouns (excluding \textit{it})
    \item \texttt{pronoun\_it}: Pronoun \textit{it}
    \item \texttt{demonstrative\_pronoun}: Demonstrative pronouns (\textit{that, this, these, those} as pronouns)
    \item \texttt{indefinite\_pronoun}: Indefinite pronouns (e.g., \textit{anybody, nothing, someone})
    \item \texttt{proverb\_do}: Pro-verb \textit{do}
\end{itemize}

\paragraph{Questions}
\begin{itemize}
    \item \texttt{wh\_question}: Direct wh-questions
\end{itemize}

\paragraph{Nominal forms}
\begin{itemize}
    \item \texttt{nominalization}: Nominalizations (ending in \textit{-tion, -ment, -ness, -ity})
    \item \texttt{gerunds}: Gerunds (participial forms functioning as nouns)
    \item \texttt{other\_nouns}: Total other nouns
\end{itemize}

\paragraph{Passives}
\begin{itemize}
    \item \texttt{agentless\_passives}: Agentless passives
    \item \texttt{by\_passives}: \textit{by}-passives
\end{itemize}

\paragraph{Stative forms}
\begin{itemize}
    \item \texttt{be\_main\_verb}: \textit{be} as main verb
    \item \texttt{existential\_there}: Existential \textit{there}
\end{itemize}

\paragraph{Subordination features}
\begin{itemize}
    \item \texttt{that\_verb\_comp}: \textit{that} verb complements (e.g., \textit{I said [that he went]})
    \item \texttt{that\_adj\_comp}: \textit{that} adjective complements (e.g., \textit{I'm glad [that you like it]})
    \item \texttt{wh\_clause}: \textit{wh}-clauses (e.g., \textit{I believed [what he told me]})
    \item \texttt{infinitives}: Infinitives
    \item \texttt{present\_participle}: Present participial adverbial clauses (e.g., \textit{[Stuffing his mouth with cookies], Joe ran out the door.})
    \item \texttt{past\_participle}: Past participial adverbial clauses (e.g., \textit{[Built in a single week], the house would stand for fifty years.})
    \item \texttt{past\_participle\_whiz}: Past participial postnominal (reduced relative) clauses (e.g., \textit{the solution [produced by this process]})
    \item \texttt{present\_participle\_whiz}: Present participial postnominal (reduced relative) clauses (e.g., \textit{the event [causing this decline]})
    \item \texttt{that\_subj}: \textit{that} relative clauses on subject position (e.g., \textit{the dog [that bit me]})
    \item \texttt{that\_obj}: \textit{that} relative clauses on object position (e.g., \textit{the dog [that I saw]})
    \item \texttt{wh\_subj}: \textit{wh}-relatives on subject position (e.g., \textit{the man [who likes popcorn]})
    \item \texttt{wh\_obj}: \textit{wh}-relatives on object position (e.g., \textit{the man [who Sally likes]})
    \item \texttt{pied\_piping}: Pied-piping relative clauses (e.g., \textit{the manner [in which he was told]})
    \item \texttt{sentence\_relatives}: Sentence relatives (e.g., \textit{Bob likes fried mangoes, [which is disgusting].})
    \item \texttt{because}: Causative adverbial subordinator (\textit{because})
    \item \texttt{though}: Concessive adverbial subordinators (\textit{although, though})
    \item \texttt{if}: Conditional adverbial subordinators (\textit{if, unless})
    \item \texttt{other\_adv\_sub}: Other adverbial subordinators (e.g., \textit{since, while, whereas})
\end{itemize}

\paragraph{Prepositional phrases, adjectives and adverbs}
\begin{itemize}
    \item \texttt{prepositions}: Total prepositional phrases
    \item \texttt{adj\_attr}: Attributive adjectives (e.g., \textit{the [big] horse})
    \item \texttt{adj\_pred}: Predicative adjectives (e.g., \textit{The horse is [big].})
    \item \texttt{adverbs}: Total adverbs
\end{itemize}

\paragraph{Lexical specificity}
\begin{itemize}
    \item \texttt{type\_token}: Type–token ratio (including punctuation)
    \item \texttt{mean\_word\_length}: Average word length (excluding punctuation)
\end{itemize}

\paragraph{Lexical classes}
\begin{itemize}
    \item \texttt{conjuncts}: Conjuncts (e.g., \textit{consequently, furthermore, however})
    \item \texttt{downtoners}: Downtoners (e.g., \textit{barely, nearly, slightly})
    \item \texttt{hedges}: Hedges (e.g., \textit{at about, something like, almost})
    \item \texttt{amplifiers}: Amplifiers (e.g., \textit{absolutely, extremely, perfectly})
    \item \texttt{emphatics}: Emphatics (e.g., \textit{a lot, for sure, really})
    \item \texttt{discourse\_particles}: Discourse particles (e.g., \textit{well, now, anyway})
    \item \texttt{demonstratives}: Demonstratives
\end{itemize}

\paragraph{Modals}
\begin{itemize}
    \item \texttt{modal\_possibility}: Possibility modals (\textit{can, may, might, could})
    \item \texttt{modal\_necessity}: Necessity modals (\textit{ought, should, must})
    \item \texttt{modal\_predictive}: Predictive modals (\textit{will, would, shall})
\end{itemize}

\paragraph{Specialized verb classes}
\begin{itemize}
    \item \texttt{verb\_public}: Public verbs (e.g., \textit{assert, declare, mention})
    \item \texttt{verb\_private}: Private verbs (e.g., \textit{assume, believe, know})
    \item \texttt{verb\_suasive}: Suasive verbs (e.g., \textit{command, propose, insist})
    \item \texttt{verb\_seem}: \textit{seem} and \textit{appear}
\end{itemize}

\paragraph{Reduced forms and dispreferred structures}
\begin{itemize}
    \item \texttt{contractions}: Contractions
    \item \texttt{that\_deletion}: Subordinator \textit{that} deletion (e.g., \textit{I think [he went]})
    \item \texttt{stranded\_preposition}: Stranded prepositions (e.g., \textit{the candidate that I was thinking [of]})
    \item \texttt{split\_infinitve}: Split infinitives (e.g., \textit{He wants [to convincingly prove] that …})
    \item \texttt{split\_auxiliary}: Split auxiliaries (e.g., \textit{They [were apparently shown] to …})
\end{itemize}

\paragraph{Co-ordination}
\begin{itemize}
    \item \texttt{phrasal\_coordination}: Phrasal coordination (\textit{N and N; Adj and Adj; V and V; Adv and Adv})
    \item \texttt{clausal\_coordination}: Independent clause coordination (\textit{clause-initial and})
\end{itemize}

\paragraph{Negation}
\begin{itemize}
    \item \texttt{neg\_synthetic}: Synthetic negation (e.g., \textit{No answer is good enough for Jones.})
    \item \texttt{neg\_analytic}: Analytic negation (e.g., \textit{That isn’t good enough.})
\end{itemize}

\section{Extended Results}
\label{app:extended-results}

\begin{figure*}[t]
    \centering
    \includegraphics[width=\linewidth]{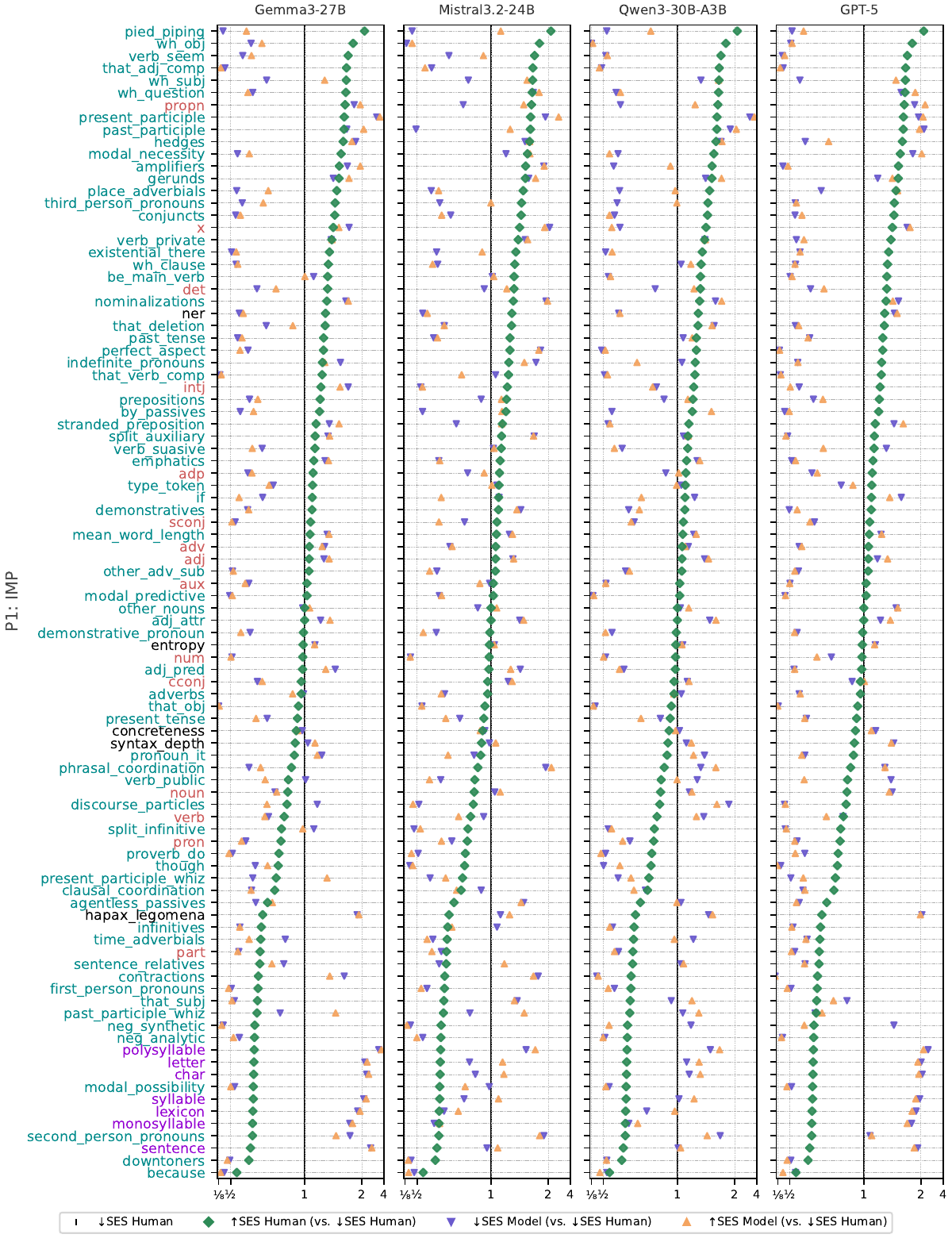}
    \caption{\textbf{Forest Plots Comparing Linguistic Features of Humans and Models on Reddit; Prompt 1, Full Features.} The plots display linguistic features between lower SES (↓SES; rate $=1$) and upper SES (↑SES) human writers. Each point indicates the frequency ratio of a feature in a model's (or human's) output compared to human text from the \lses{} group. These comparisons are shown across four models and three prompts (see Section~\ref{sec:prompts}). Feature types are color-coded: Biber features are cyan, length-specific features are dark violet, PoS tags are red, and style features are black.}
    \label{fig:reddit1}
\end{figure*}

\begin{figure*}[t]
    \centering
    \includegraphics[width=\linewidth]{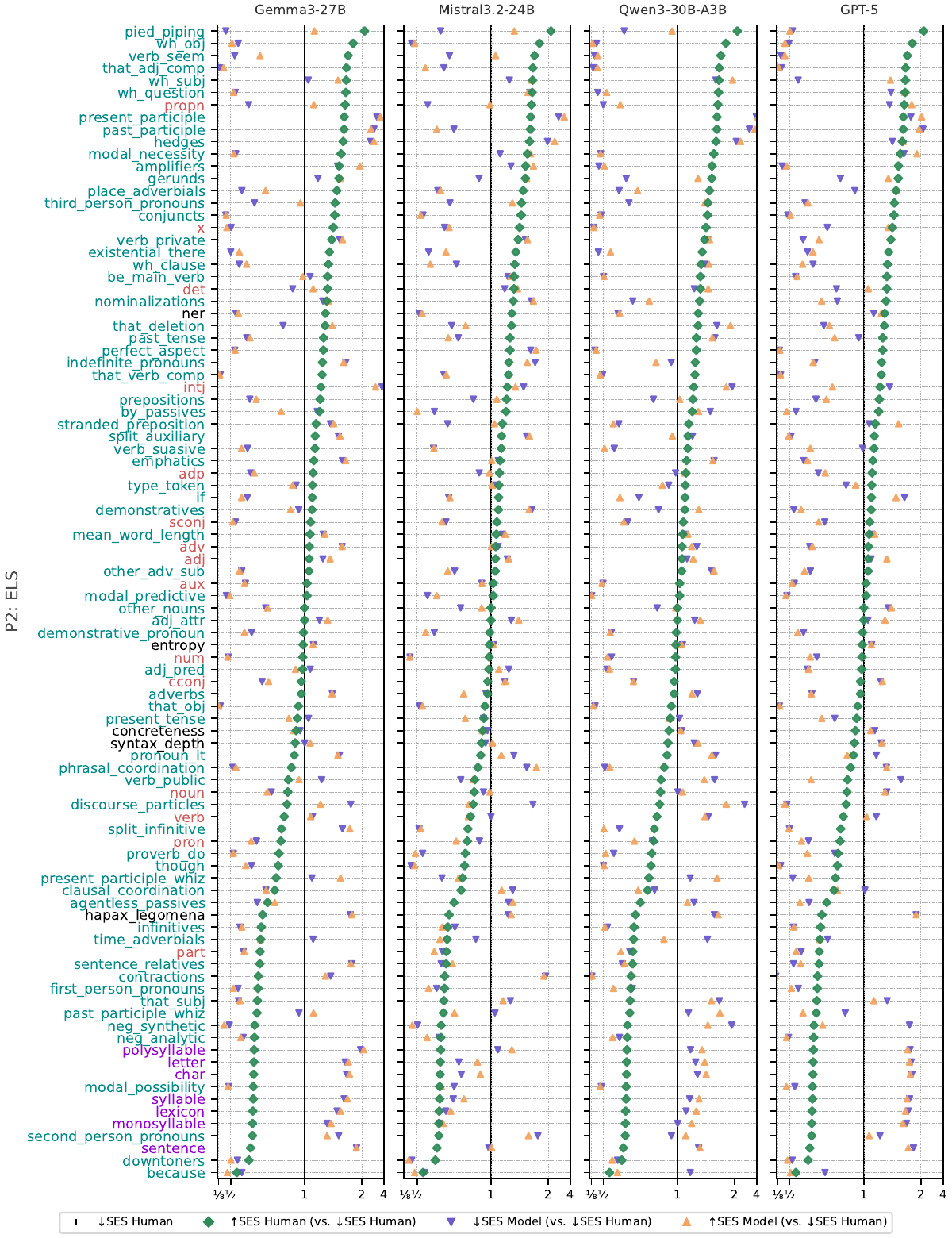}
    \caption{\textbf{Forest Plots Comparing Linguistic Features of Humans and Models on Reddit; Prompt 2, Full Features.} The plots display linguistic features between lower SES (↓SES; rate $=1$) and upper SES (↑SES) human writers. Each point indicates the frequency ratio of a feature in a model's (or human's) output compared to human text from the \lses{} group. These comparisons are shown across four models and three prompts (see Section~\ref{sec:prompts}). Feature types are color-coded: Biber features are cyan, length-specific features are dark violet, PoS tags are red, and style features are black.}
    \label{fig:reddit2}
\end{figure*}

\begin{figure*}[t]
    \centering
    \includegraphics[width=\linewidth]{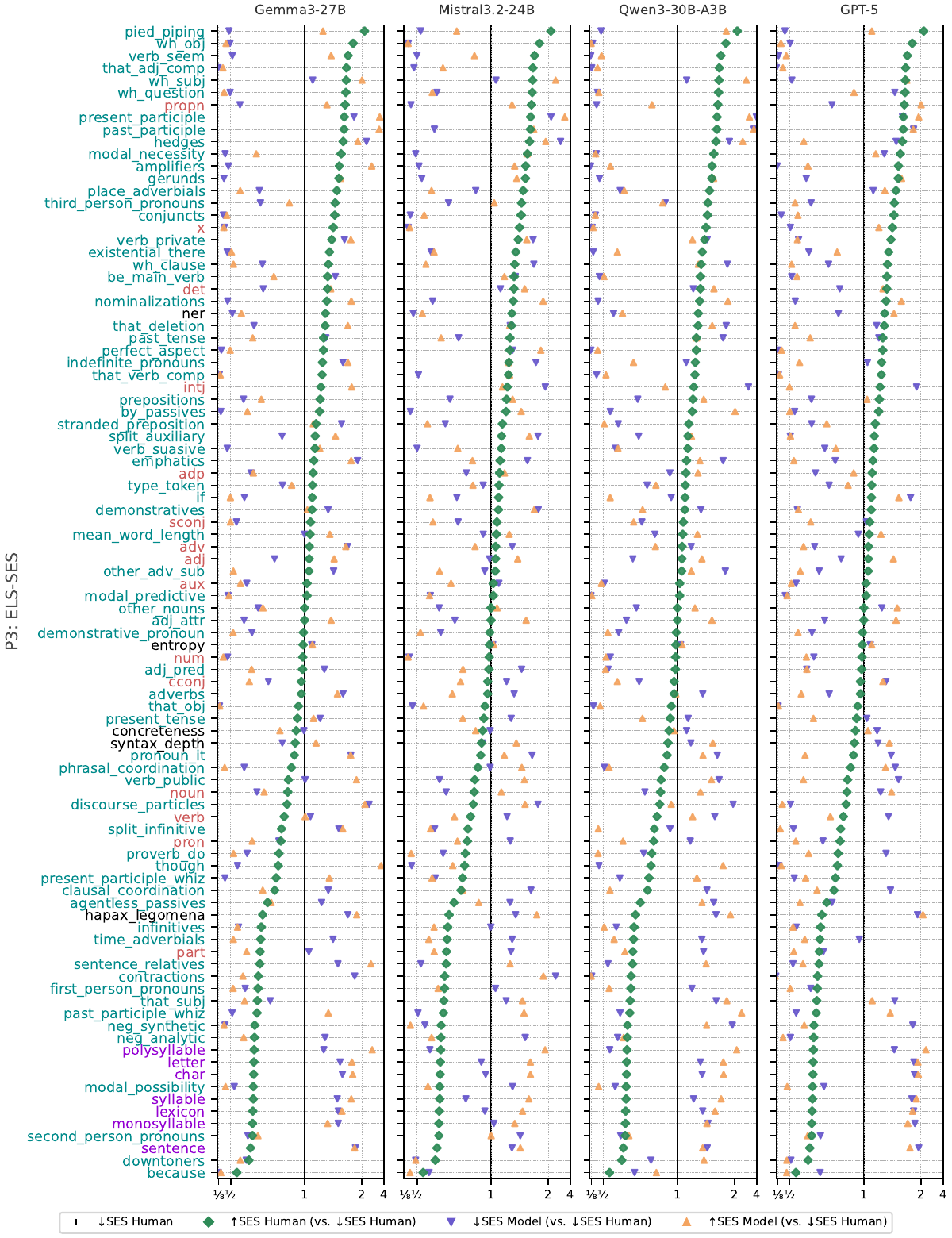}
    \caption{\textbf{Forest Plots Comparing Linguistic Features of Humans and Models on Reddit; Prompt 3, Full Features.} The plots display linguistic features between lower SES (↓SES; rate $=1$) and upper SES (↑SES) human writers. Each point indicates the frequency ratio of a feature in a model's (or human's) output compared to human text from the \lses{} group. These comparisons are shown across four models and three prompts (see Section~\ref{sec:prompts}). Feature types are color-coded: Biber features are cyan, length-specific features are dark violet, PoS tags are red, and style features are black.}
    \label{fig:reddit3}
\end{figure*}

\begin{figure*}[t]
    \centering
    \includegraphics[width=\linewidth]{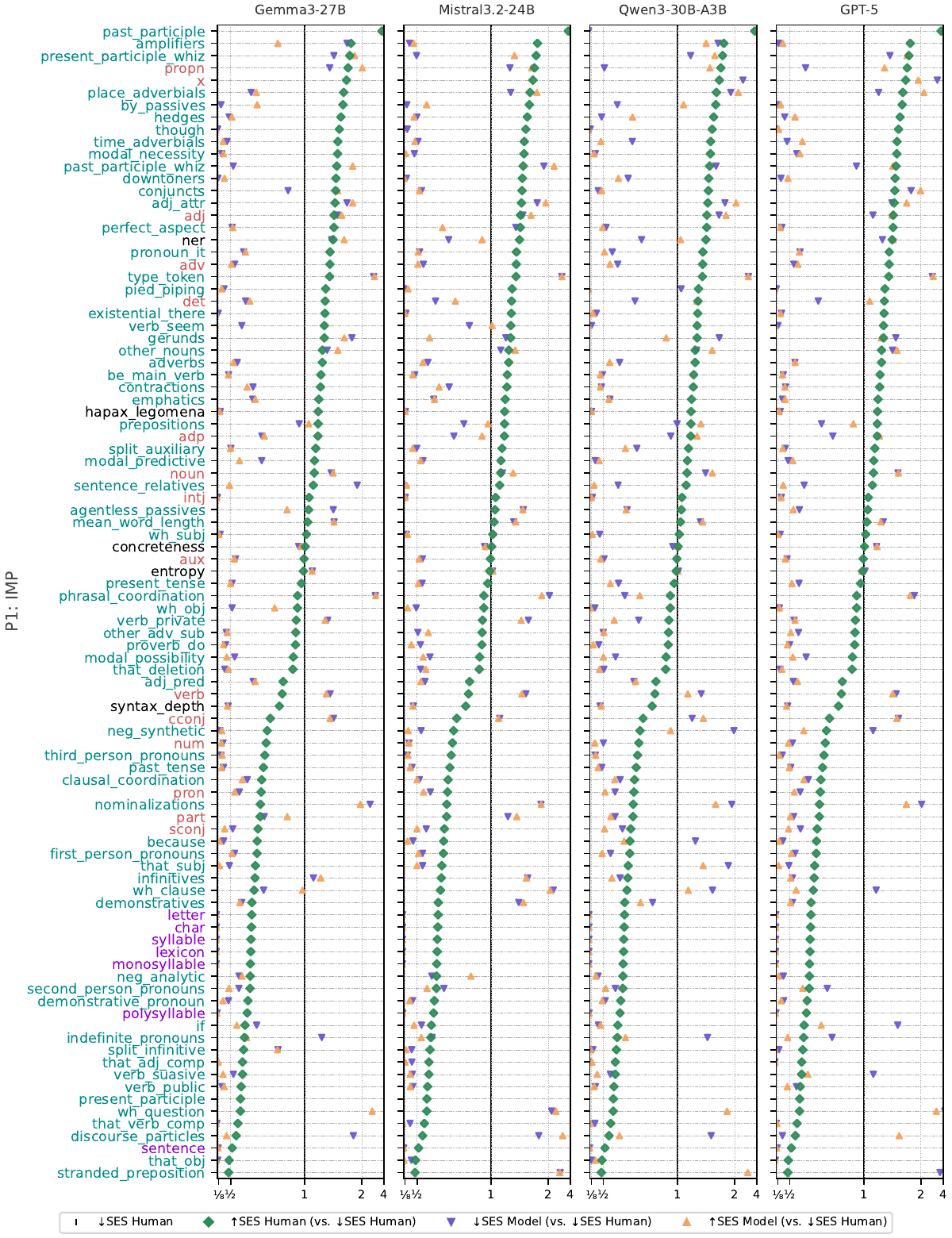}
    \caption{\textbf{Forest Plots Comparing Linguistic Features of Humans and Models on Youtube; Prompt 1, Full Features.} The plots display linguistic features between lower SES (↓SES; rate $=1$) and upper SES (↑SES) human writers. Each point indicates the frequency ratio of a feature in a model's (or human's) output compared to human text from the \lses{} group. These comparisons are shown across four models and three prompts (see Section~\ref{sec:prompts}). Feature types are color-coded: Biber features are cyan, length-specific features are dark violet, PoS tags are red, and style features are black.}
    \label{fig:yt1}
\end{figure*}

\begin{figure*}[t]
    \centering
    \includegraphics[width=\linewidth]{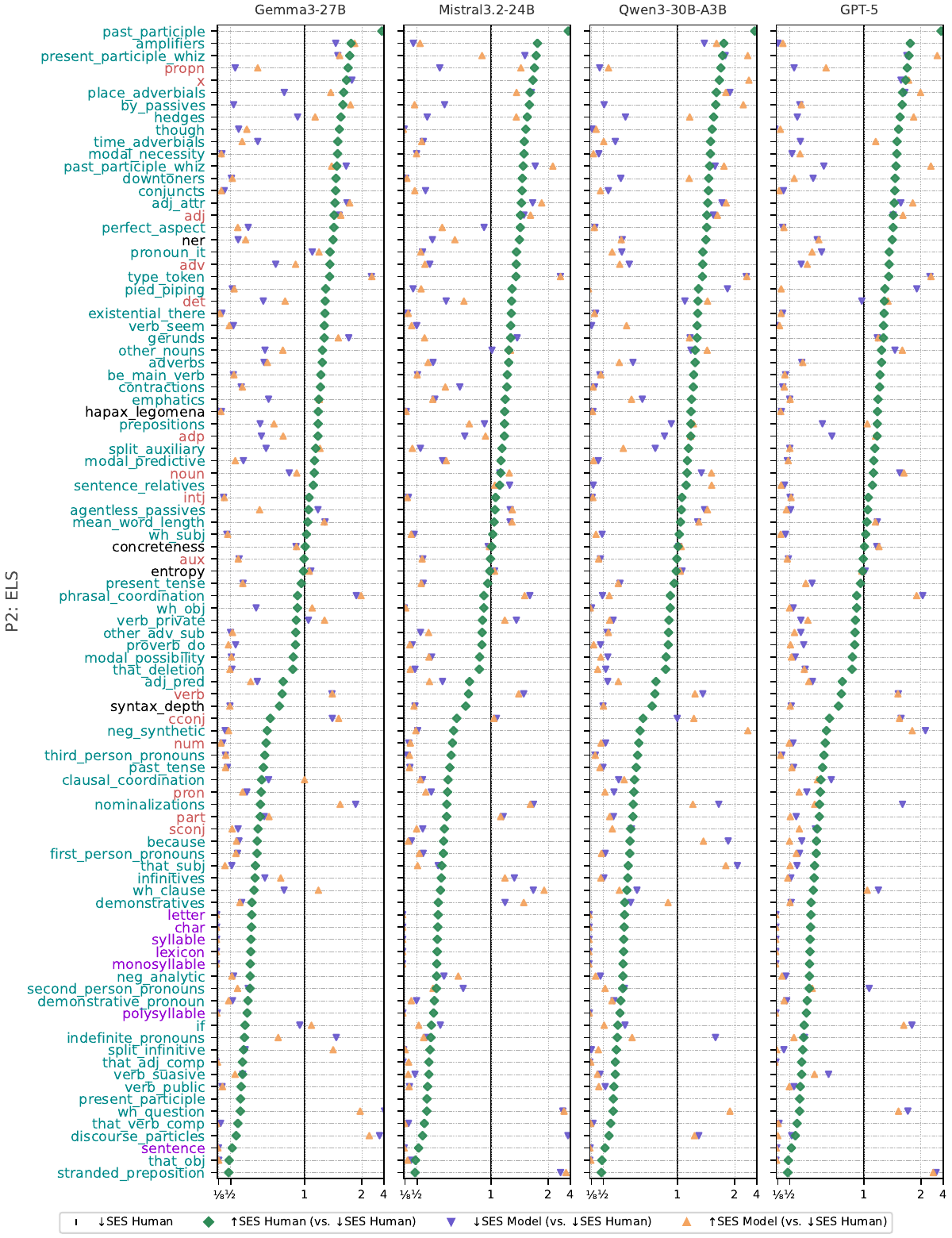}
    \caption{\textbf{Forest Plots Comparing Linguistic Features of Humans and Models on Youtube; Prompt 2, Full Features.} The plots display linguistic features between lower SES (↓SES; rate $=1$) and upper SES (↑SES) human writers. Each point indicates the frequency ratio of a feature in a model's (or human's) output compared to human text from the \lses{} group. These comparisons are shown across four models and three prompts (see Section~\ref{sec:prompts}). Feature types are color-coded: Biber features are cyan, length-specific features are dark violet, PoS tags are red, and style features are black.}
    \label{fig:yt2}
\end{figure*}

\begin{figure*}[t]
    \centering
    \includegraphics[width=\linewidth]{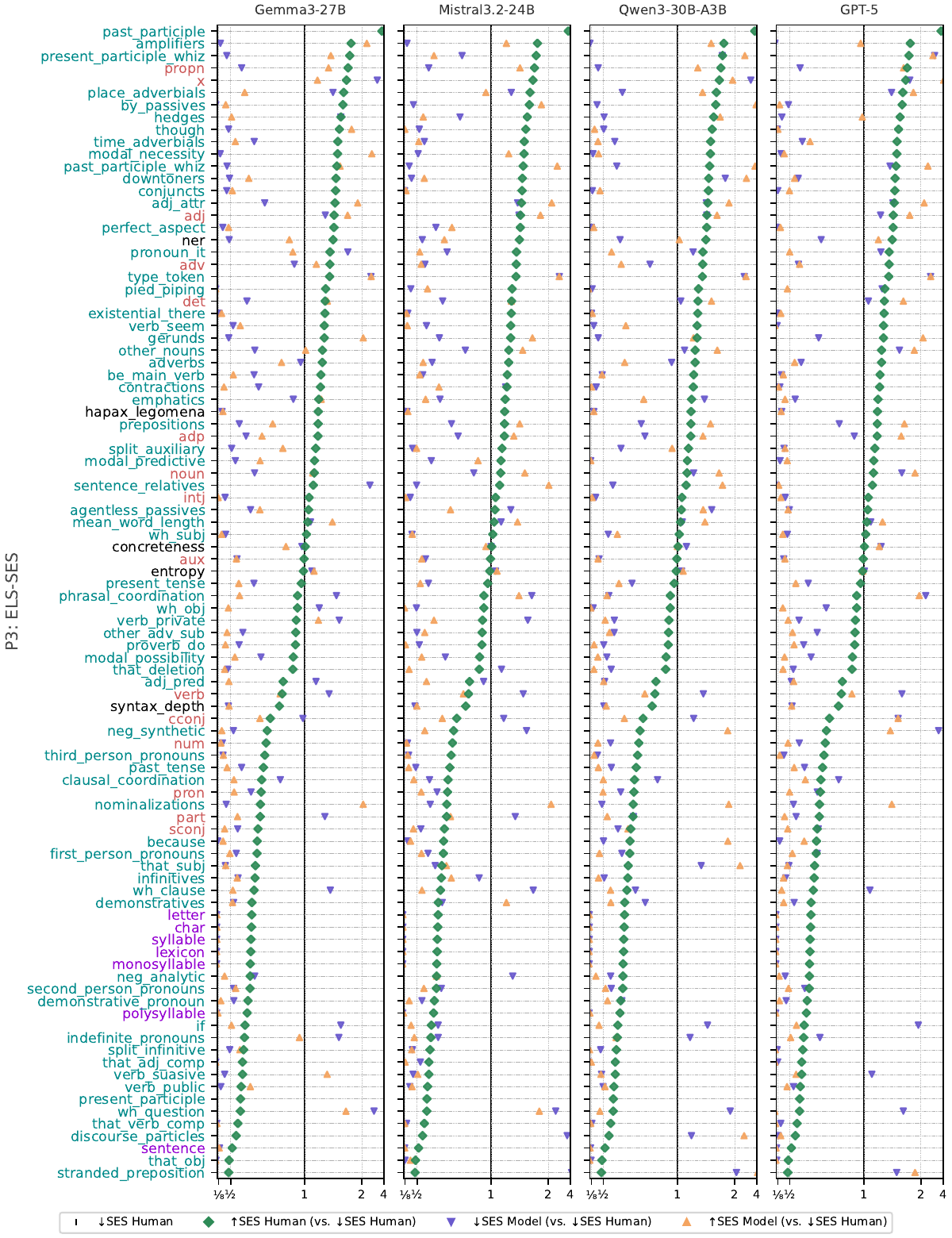}
    \caption{\textbf{Forest Plots Comparing Linguistic Features of Humans and Models on Youtube; Prompt 3, Full Features.} The plots display linguistic features between lower SES (↓SES; rate $=1$) and upper SES (↑SES) human writers. Each point indicates the frequency ratio of a feature in a model's (or human's) output compared to human text from the \lses{} group. These comparisons are shown across four models and three prompts (see Section~\ref{sec:prompts}). Feature types are color-coded: Biber features are cyan, length-specific features are dark violet, PoS tags are red, and style features are black.}
    \label{fig:yt3}
\end{figure*}

In~\Cref{fig:reddit1},~\Cref{fig:reddit2}, and~\Cref{fig:reddit3}, we show the full feature list results for Reddit. Moreover, in~\Cref{fig:yt1},~\Cref{fig:yt2}, and~\Cref{fig:yt3}, we show the full feature list results for YouTube.

\section{Inference Experiments}\label{app:inference}
For running \texttt{GPT-5}, we use the OpenAI\footnote{\url{https://platform.openai.com/}} API. The costs of running inference on all data took around 112 USD. For running inference of the local models (Gemma3, Mistral, Qwen3), we make use of a large HPC cluster with hardware configurations comprising multiple nodes (depending on model size; e.g., 30B models require 4 nodes for training and 1 node for inference), each with node contains eight AMD MI250x GPU modules alongside a single 64-core AMD EPYC ``Trento'' CPU. The library we use for inference is \texttt{vllm}~\citep{kwon2023efficient}. For all the experiments it resulted in around 16 GPU hours spent. 

\subsection{Environmental Impact}
We acknowledge that conducting a large-scale analysis using LLMs comes with an environmental impact. Experiments were conducted using private infrastructure in \texttt{[Redacted]} running on green energy. A cumulative of 16 GPU hours of computation was performed on AMD MI250x GPU modules, which has a TDP of 500 Watts. The experiments were ran in September 2025. During this time, the average carbon efficiency in \texttt{[Redacted]} was 0.046 $kg/kWh$.\footnote{According to \url{https://app.electricitymaps.com/map}.} This means we released about 0.368 $kg$ of $CO_2$ equivalent. Estimations were conducted using the Machine Learning Impact calculator\footnote{Find the tool here: \url{https://mlco2.github.io/impact}.} presented in \citep{lacoste2019quantifying}.

\end{document}